\pdfoutput=1

\documentclass[11pt]{article}

\usepackage{acl}

\usepackage{times}
\usepackage{latexsym}

\usepackage[T1]{fontenc}

\usepackage[utf8]{inputenc}

\usepackage{microtype}
\usepackage{adjustbox}

\usepackage{pdfpages} 

\usepackage{graphicx}
\usepackage{tabularx}
\usepackage{soul}
\usepackage{CJKutf8}
\usepackage{booktabs, multirow} 
\usepackage{soul}

\usepackage{arydshln}
\usepackage{pgfplots}

\usepackage[caption=false]{subfig}

\title{NUMCoT: Numerals and Units of Measurement in Chain-of-Thought Reasoning using Large Language Models}

\newcommand*\samethanks[1][\value{footnote}]{\footnotemark[#1]}

\newcolumntype{Y}{>{\centering\arraybackslash}X}

\author{
Ancheng Xu$^{1,2}$ \quad Minghuan Tan$^{1}$\thanks{~~Corresponding author.} \quad Lei Wang$^{3}$ \quad Min Yang$^{1}$\samethanks[1] \quad Ruifeng Xu$^{4}$ \\
$^{1}$Shenzhen Institute of Advanced Technology, Chinese Academy of Sciences \\
$^{2}$ University of Chinese Academy of Sciences\\
$^{3}$ School of Computing and Information Systems, Singapore Management University\\
$^{4}$ Harbin Institute of Technology (Shenzhen)\\
\{ac.xu,mh.tan,min.yang\}@siat.ac.cn, lei.wang.2019@phdcs.smu.edu.sg, xuruifeng@hit.edu.cn
}

\begin{document}
\maketitle

\begin{CJK*}{UTF8}{gbsn}

\begin{abstract}
Numeral systems and units of measurement are two conjoined topics in activities of human beings and have mutual effects with the languages expressing them.
Currently, the evaluation of Large Language Models (LLMs) often involves mathematical reasoning, yet little attention is given to how minor changes in numbers or units can drastically alter the complexity of problems and the performance of LLMs.
In this paper, we scrutinize existing LLMs on processing of numerals and units of measurement by constructing datasets with perturbations.
We first anatomize the reasoning of math word problems to different sub-procedures like numeral conversions from language to numbers and measurement conversions based on units.
Then we further annotate math word problems from ancient Chinese arithmetic works 
which are challenging in numerals and units of measurement.
Experiments on perturbed datasets demonstrate that LLMs still encounter difficulties 
in handling numeral and measurement conversions.
The code and data are available at: \href{https://github.com/CAS-SIAT-ConsistencyAI/NUMCoT}{https://github.com/CAS-SIAT-ConsistencyAI/NUMCoT}.
\end{abstract}

\section{Introduction}
\label{sec:intro}
Numbers and counting are the basic concepts in human experience.
Numbers are a set of conceptual tools made from words and other symbols for specific quantities and a key set of linguistically based innovations that distinguish human species from others~\cite{everett2017numbers}.
The development of numeral systems allows humans to express numbers in a consistent manner.\footnote{\url{https://en.wikipedia.org/w/index.php?title=Numeral_system}}
Counting is usually not a monotone process of manipulating numbers from a numeral system but to quantify objects with a units of measurement\footnote{\url{https://en.wikipedia.org/w/index.php?title=Unit_of_measurement}} to compare the magnitude.

In the literature, \citet{thawani-etal-2021-representing} adopt the taxonomy discipline called Core Systems of Number~\cite{FEIGENSON2004307} from cognitive science.
The tasks in numeracy are then categorized by the granularity and units attached to the quantities in the task, where granularity means whether the encoding of the number is exact or approximate, and units represent  whether the numerals are in their numerical forms or grounded with units of measurement. 
Based on the taxonomy, existing numeracy-oriented tasks are identified as simple arithmetic tasks~\cite{generalization}, numeration tasks~\cite{naik-etal-2019-exploring,wallace-etal-2019-nlp,johnson-etal-2020-probing}, magnitude comparison tasks~\cite{naik-etal-2019-exploring,wallace-etal-2019-nlp}, Math Word Problems (MWPs)~\cite{roy-roth-2015-solving,wang-etal-2017-deep,amini-etal-2019-mathqa}, exact facts in the context of numeracy~\cite{lin-etal-2020-birds,mishra2020question}, measurement estimation tasks~\cite{forbes-choi-2017-verb,elazar-etal-2019-large,zhou-etal-2020-temporal} and numerical language modeling tasks.
There are still tasks which fall out the taxonomy, such as numeric paraphrasing (one-to-one correspondences between different surface forms of the same number), quantity entailment tasks~\cite{mishra2020question}, numeral understanding tasks, Fused-Head Resolution, counting tasks~\cite{suzgun-etal-2019-lstm,bhattamishra-etal-2020-computational} and other domain-specific tasks. 
As far as we are concerned, the tasks discussed above cover a wide range of topics in numeracy and address a lot of challenges faced by numerals and units of measurement.

\begin{figure*}[!hpt]
\centering
\includegraphics[width=\linewidth]{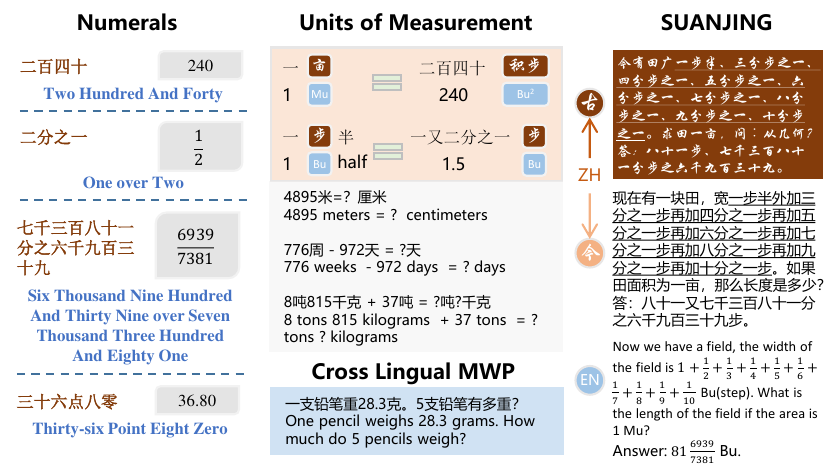}
\caption{On the left of the image are numeral conversions tasks. In the middle are challenges related to unit conversion and mathematical problems. On the far right is an example from \textsc{SuanJing}, featuring its original problem in ancient Chinese.} 
\label{fig:datasets}
\end{figure*}

However, we still need to address the issue of numeracy when discussing arithmetic by using pure numerals and making an extra effort to take units of measurement into consideration.
The inadequacy in accurately converting numerals with units of measurement may lead to unpredictable consequences in real-life scenarios, especially in the era of Large Language Models (LLMs) where decoder-only generation methods are being employed.

Conventional LLMs~\cite{workshop2023bloom,openai2023gpt4,zeng2023glmb,touvron2023llama}  implicitly assume that numeral systems and units of measurement are innate, and they conduct analysis at the reasoning level to demonstrate their ability in solving math word problems.
For example, \citet{wei2023chainofthought} uses Chain-of-Thought method to prompt LLMs to generate coherent series of intermediate reasoning steps to solve problems.
We argue that this assumption needs further verification, and better prompting methods are also needed to explore the extent to which the assumption actually works. 
In a math word problem, if the conversion of numerals or scale of units fails, it's not guaranteed to be correct, even if each further reasoning step is on the right track.
We justify our claim from the following aspects:
(1) In LLMs, the extrapolation of numerals is more difficult to define, as the numbers in the training set have a wider range compared to traditional models.
(2) Although most math word problems adopt the Hindu-Arabic writing style for numerals, it's still common to use pronunciations with written style to mark a number for the advantage of being irrevocable, especially in Chinese.
When writing Arabic numerals, we often overlook the magnitude and only focus on the length of the numerals.
However, the rules for reading numbers are very different. For example, in English, every three digits are divided into a scale, while in Chinese, it is every four digits.
When pronouncing, we first focus on the length of the numbers, then on the magnitude, and finally group and read them one by one.
(3) To our best knowledge, the investigation on units of measurement has been conducted through measuring skill tests (unit conversion,
reference range detection, and measure comparison)~\cite{park-etal-2022-language} with pretrained language models and has identified their lack of such abilities.
It is still unknown to what extent LLMs can overcome this challenge, especially in uniting numeral conversions with units of measurement.

To achieve these goals, we construct four datasets to synthesize the procedure of how humans process numerals and units of measurement.
The procedure is anatomized into sub-procedures like converting words into numbers, dealing with units of measurement with different scales and solving the problem using reasoning and rationale.
For each sub-procedure, we employ random numbers and addition operations to perturb the dataset, thereby reducing the generation of memorization issues.

In this paper, we focus on ChatGPT~\cite{openai-chatgpt-2022}, ChatGLM series models~\cite{zhipu2023chatglm} , ERNIE-Bot~\cite{baidu2023erniebot} and LLaMA-2 family models~\cite{touvron2023llama2}. We construct different prompts to elicit LLMs to generate responses for the datasets above. Our experiments reveal that LLMs exhibit robustness in converting between numbers and English text, but less effectiveness in converting between numbers and Chinese text.

Furthermore, LLMs consistently struggle to memorize conversion ratios between different units, posing challenges for automatic numeral conversions based on unit changes. In MWPs involving numeral conversions and units of measurement, LLMs perform well. However, LLMs often struggle to provide correct answers to \textsc{SuanJing} problems that require specialized long-tail knowledge.

In summary, our work makes the following contributions:
\begin{enumerate}
    \item We construct four datasets to explore the performance of LLMs in tasks that involve numeral conversions and unit conversion, which are crucial research questions for LLMs.
    \item We discover and verify that introducing CoT in certain subtasks significantly deteriorates the reasoning performance of LLMs. In the experimental section, we provide the corresponding analysis.
    \item We conduct prompt-based experiments on LLMs to assess their ability in numeral conversions and units of measurement, thereby highlighting a new direction for training and benchmarking LLMs.
\end{enumerate}

\section{Related Work}

\subsection{Units of Measurement in Numeracy}

Units of measurement in numeracy have been attracting attention from the community because of their relationship with common sense in life and domain knowledge in applications. Despite recent success of pre-trained language models~(PLMs)~\cite{devlin-etal-2019-bert,liu2020roberta}, their reasoning abilities using numerical commonsense is surprisingly poor~\cite{lin-etal-2020-birds} and PLMs lack the capability required for reasoning over measurements~\cite{park-etal-2022-language}. 
The knowledge on scaling of measurement, such as \textit{1000 meters make a km}, can add extra challenge to numerical reasoning tasks~\cite{mishra-etal-2022-numglue}.

While traditional explorations over measurements address more on quantity identification with measurements~\cite{harper-etal-2021-semeval,gopfert-etal-2022-measurement} and their comparable properties~\cite{forbes-choi-2017-verb,lin-etal-2020-birds,park-etal-2022-language}, we focus more on the accuracy of their usage from arithmetic perspective.
With the development of CoT-based approaches in LLMs, we are also curious how they perform on dealing with different system of units in either base forms and derived forms.

\subsection{Numeracy in Large Language Models}

Besides the survey conducted by \citet{thawani-etal-2021-representing} that is mentioned in Section~\ref{sec:intro}, we also review how numeracy is discussed in the era of LLMs.
The evaluation of GPT-3~\cite{brown2020language} over NumGLUE~\cite{mishra-etal-2022-numglue} indicates that it is a better few-shot learner but not necessarily a better many-shot learner.
In arithmetic, MathGLM~\cite{yang2023gpt} breaks the misconception that LLMs are unable to accurately perform arithmetic operations and trains a model which can accurately perform multi-digit arithmetic operations with almost 100\% accuracy without data leakage, significantly surpassing GPT-4~\cite{openai2023gpt4}. 

\section{Datasets and Perturbations}

\subsection{Datasets}

For math word problems using different numeral systems and units of measurement, we are curious about how LLMs process such information in their reasoning steps.
We choose to anatomize the reasoning of math word problems into different sub-procedures, like conversions between numbers and words, conversions with units of measurement.
We first build the \textit{Numeral Conversions} dataset and the \textit{Conversions with Units of Measurement} dataset.
Then we construct the \textit{Cross Lingual MWPs} dataset that involves math word problems with Chinese and English, and the \textsc{SuanJing} dataset abundant with these challenges.
The datasets are illustrated in Figure~\ref{fig:datasets}.

\paragraph{Numeral Conversions} The conversion of numerals to words (\textit{Num2Words}) and its inverse process ~\textit{Words2Num} are two basic abilities for humans to manipulate numbers.
Pronunciation of numerals is critical for humans to express quantities precisely.
For example, an integer 21,600,900 should be pronounced as ``twenty one million six hundred thousand nine hundred only" in English and ``二千一百六十万零九百" in Chinese.
The task is also called as (Numeric) Paraphrasing~\cite{thawani-etal-2021-representing}.
The practice using text conversion from numerical to standard spelled-out numbers in numeracy probing has been conducted earlier in other multilingual numerical understanding works~\cite{johnson-etal-2020-probing}.

Different from them, where numbers are generated from a smaller range of 0 to 999, we generate numbers from 0 to trillions and consider the complexity of each number from both scale and pronunciation forms.
The ~\textit{Numeral Conversions} dataset is separated into the following splits:
\begin{enumerate}
    \item The \textit{Numeral Conversions Medium} split consists of 400 randomly generated integers falling into the ranges of zero to a thousand~(0-1K), a thousand to a million~(1K-1M), a million to a billion~(1M-1B), and a billion to a trillion~(1B-1T), with each range containing 100 integers.
    \item The \textit{Numeral Conversions Easy} split comprises 400 Arabic numerals with lengths identical to those in the \textit{Numeral Conversions Medium} split, but the corresponding pronunciation forms in Chinese and English are significantly shorter.
    \item The \textit{Numeral Conversions Hard} split consists of 200 fractions and 200 decimals. For fractions, the numerators and denominators of the fractions are randomly sampled from the same four numerical ranges mentioned earlier, ensuring they are of similar scales. Two random integers, A and B, are generated within their respective numerical range, forming a fraction in the format \textbf{A/B} . Similarly, two random integers, C and D, are selected within their corresponding numerical range, composing a decimal in the format \textbf{C.D} .
\end{enumerate}

\paragraph{Conversions with Units of Measurement} In most human experiences, numbers are used in joint with units of measurement to express real-world quantities. In specific scenarios, units of measurement with different scales are also ubiquitous.
For example, 1.5 litre is equivalent to 1 litre plus 500 milliliters.
However, it's still questionable whether LLMs process such information similarly as humans.

To emphasize this sub-procedure, we create parallel datasets in both Chinese and English based on 18 units commonly used by humans, such as length, time, weight, and money, including centimeters, seconds, kilograms, yuan, and other units.
These datasets are generated using random numbers and are identical in all aspects except for the language. 
Additionally, we categorize the questions into three levels of difficulty.
\begin{enumerate}
    \item The \textit{Units of Measurement Easy} split involves the conversion of numerical values from one unit to another. For example, 856 grams = ? milligrams.
    \item The \textit{Units of Measurement Medium} split requires performing addition or subtraction between two units before converting to another unit. For example, 738 seconds  - 5 milliseconds  = ? milliseconds.
    \item The \textit{Units of Measurement Hard} split involves a more complex process: combining two units into one and then performing addition or subtraction operations before converting to another unit. For example, 4 days 387 hours  + 81 days  = ? days ? hours.
\end{enumerate}
LLMs require common sense and reasoning abilities to complete conversions at all three levels.

\paragraph{MWPs and \textsc{SuanJing}}To compare the challenges introduced by numeral conversions and units of measurement, we utilize a bilingual MWPs dataset redacted by \citet{tan-etal-2022-investigating} and a Chinese dataset \textsc{SuanJing} translated from ancient Chinese MWPs. The bilingual MWPs dataset is compiled from AddSub~\cite{hosseini-etal-2014-learning}, SingleOp~\cite{roy-etal-2015-reasoning} and MultiArith~\cite{roy-roth-2015-solving}, containing 1557 elementary school math word problems.

\textsc{SuanJing} problems are constructed by translating ancient Chinese to modern Chinese while preserving character-level numeral representations.
We select \textsc{SuanJing} because it comprehensively tests LLMs on tasks like Num2Words, Words2Num, and Conversations with Units of Measurement. This setup allows us to examine LLMs' performance under various conditions: without CoT, with CoT but lacking rare knowledge, and with CoT plus rare knowledge.
The translation is performed by ChatGLM-6B~\cite{du-etal-2022-glm,zeng2022glm} and further refined by human experts.
We list details about \textsc{SuanJing} in Appendix~\ref{sec:appendix}.

\begin{table*}[!hpt]
\centering
\includegraphics[width=\linewidth]{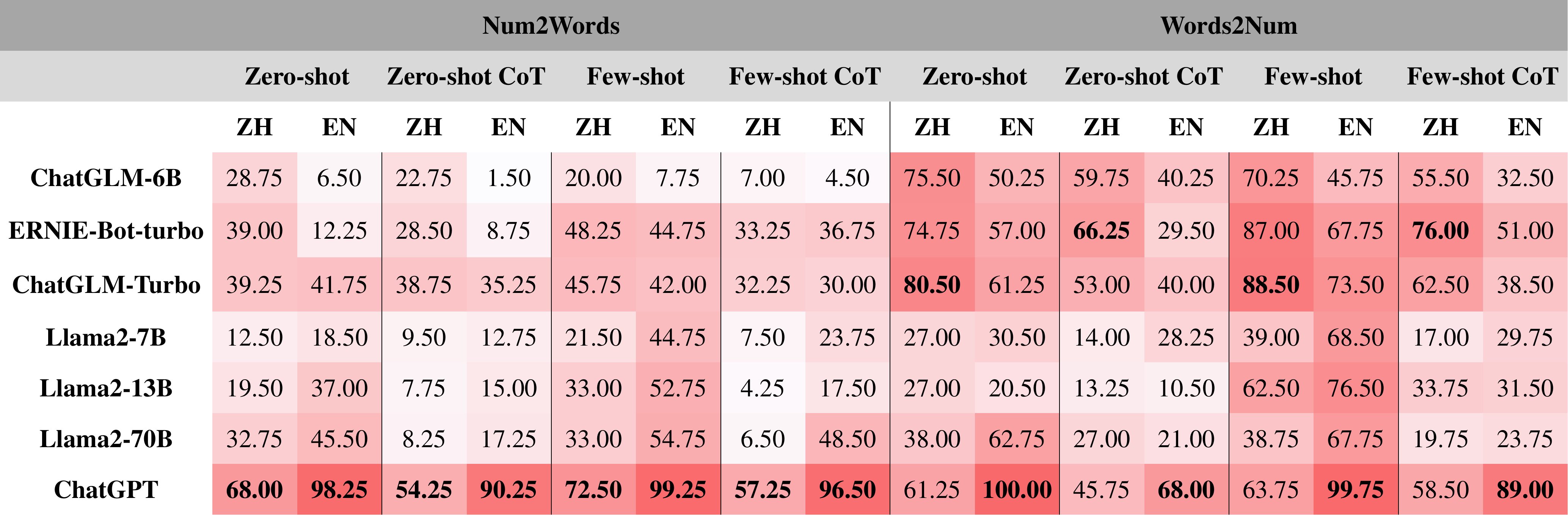}
\caption{Overview of conversion accuracy for Num2words and Words2Num on the Numeral Conversions Medium split using the four prompt methods: Zero-shot, Zero-shot with CoT, Few-shot and Few-shot with CoT.} 
\label{tab:overview1}
\end{table*}

\subsection{Perturbations}

To avoid the generation from memorization issues that might occur with LLMs, we decide to perturb the datasets created above. For example, to design a dataset with Arabic numeral lengths equal to those in the \textit{Numeral Conversions Medium} dataset, and with Chinese and English representations shorter than those in the \textit{Numeral Conversions Medium} dataset, the numerical format of the \textit{Numeral Conversions Easy} dataset should ideally follow that of  \( M \times 10^N \). 

However, considering the likelihood of LLMs encountering \textit{Numeral Conversions Easy} dataset numbers frequently during pretraining, we introduce perturbations by adding one to each number in the \textit{Numeral Conversions Easy} dataset, with the format being \(M\times 10^N + 1\).

\section{Experiments}

We conduct the experiments using open-sourced LLMs as well as API-based LLMs supporting both English and Chinese languages.
For publicly available LLMs, we chose ChatGLM2-6B\footnote{\url{https://github.com/THUDM/ChatGLM2-6B}} and three models from the LLaMA-2\footnote{\url{https://llama.meta.com/llama2}} family: 7B, 13B, and 70B, which were deployed to A6000 GPU server locally.
For API-based LLMs, we use ChatGPT\footnote{\url{https://platform.openai.com/docs/models/gpt-3-5}}, ERNIE-Bot-turbo\footnote{\url{https://cloud.baidu.com/doc/WENXINWORKSHOP/s/4lilb2lpf}}, ChatGLM-Turbo\footnote{\url{https://open.bigmodel.cn/dev/howuse/model}}.

\begin{figure*}[ht]
\centering
\subfloat[Overall performance of different models.\label{fig:radar_chart_overall}]{%
  \includegraphics[width=0.23\textwidth]{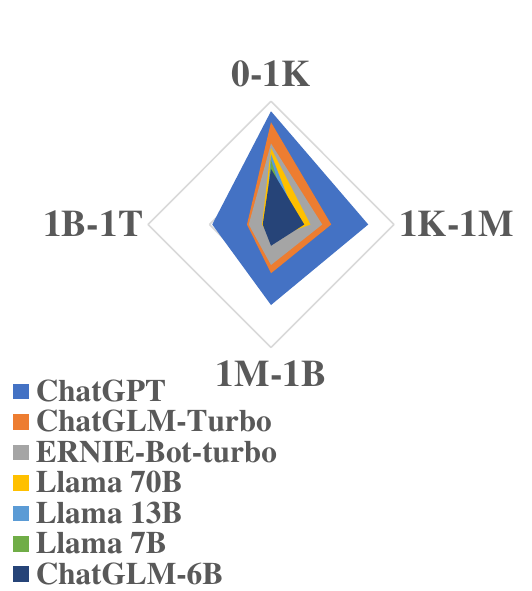}%
}\hfil
\subfloat[Accuracy of ChatGPT on splits with different complexity.\label{fig:radar_chart_complexity}]{%
  \includegraphics[width=0.23\textwidth]{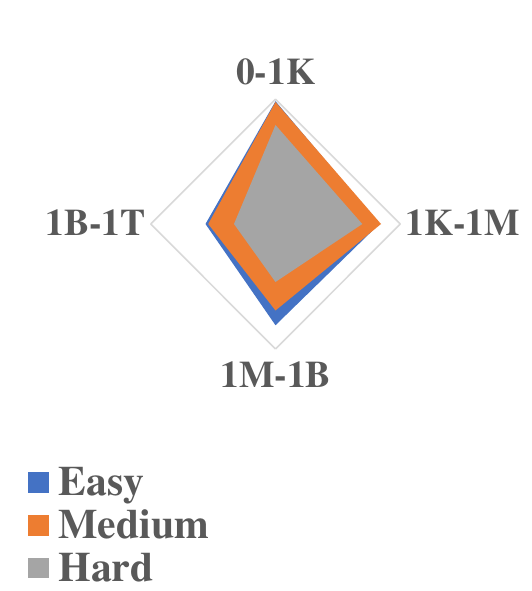}%
}\hfil
\subfloat[Accuracy of ChatGPT with different prompts.\label{fig:radar_chart_prompt}]{%
  \includegraphics[width=0.23\textwidth]{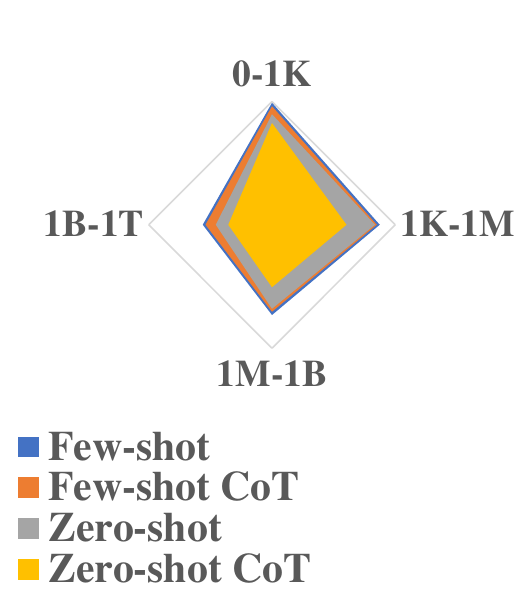}%
}\hfil
\subfloat[Accuracy of ChatGPT with different languages.\label{fig:radar_chart_language}]{%
  \includegraphics[width=0.23\textwidth]{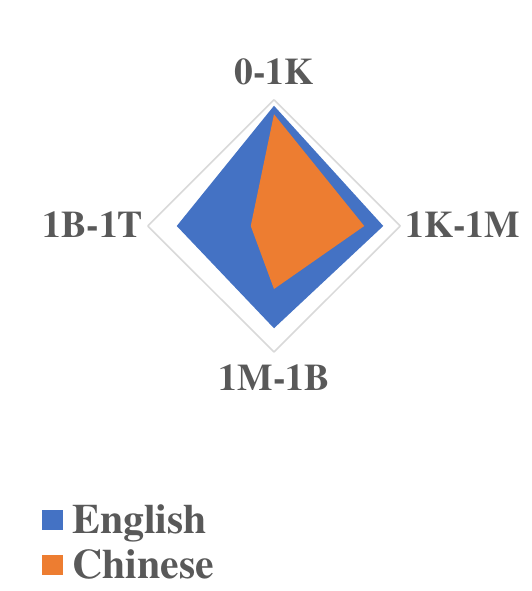}%
}
\caption{Accuracy against different scales with respect to different dimensions.} \label{fig:radar}
\end{figure*}

\begin{table*}[t]
\centering
\includegraphics[width=\linewidth]{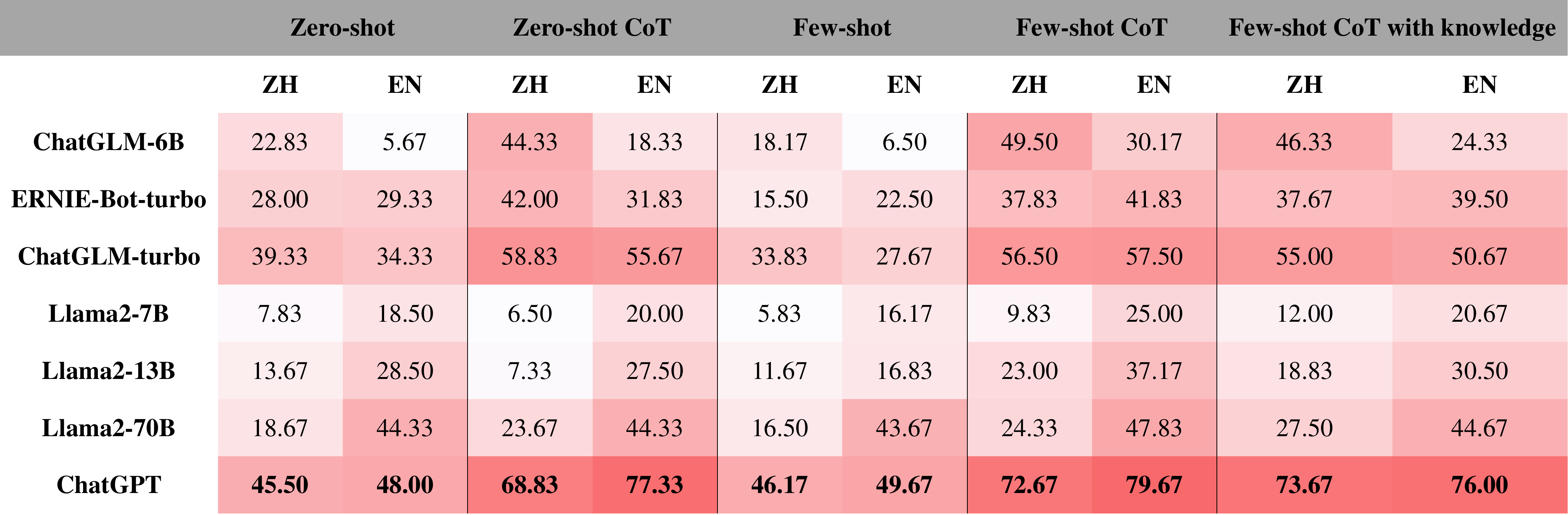}
\caption{Overview of reasoning accuracy for Units of
Measurement on the Numeral Conversions Medium split using the five prompt methods: Zero-shot, Zero-shot CoT, Few-shot, Few-shot CoT and Few-shot CoT with knowledge.}
\label{tab:unit_experience}
\end{table*}

We consider the following prompt:
(1) \textbf{Zero-shot}：We simply present the questions to the LLMs without introducing any examples, reasoning steps, or CoT. 
(2) \textbf{Zero-shot CoT}: We simply present the questions to the LLMs, employing the CoT framework without introducing any examples or deductive steps. Our approach involve the simple addition of the phrase \textit{Let's think step by step}.
(3) \textbf{Few-shot}: We present four analogous questions accompanied by concise responses in the prompt before presenting the questions to the LLMs, without introducing deductive steps.
(4) \textbf{Few-shot CoT}: We present four analogous questions, each accompanied by concise responses, within the prompt prior to presenting them to the LLMs. Additionally, deductive steps are introduced alongside the questions.

\subsection{The Accuracy of Numeral Conversions}
 
We list the experimental results for the \textit{Numeral Conversions Medium} splits in Table~\ref{tab:overview1}. For more information about the prompt design for the current experiment, please refer to Table~\ref{tab:prompt_num2word-1} to \ref{tab:prompt_word2num-3} in the Appendix.

We have the following findings:
(1) ChatGPT has significant advantages over other models in conversions using English and is almost perfect at \textit{Num2Words} task.
(2) Introducing CoT and deductive steps in the \textit{Num2Words} and \textit{Words2Num} tasks results in a significant decrease in accuracy compared to prompts without the incorporation of CoT and deductive steps.

\paragraph{Accuracy against Different Scales}

From a numerical scale perspective, different models exhibit significant variance in performance, with ChatGPT outperforming all other models. 
When the number is less than 1000, all models achieve their best performance, and the gap is smallest compared to that of ChatGPT. 
However, as the scale of numbers increases, there is a consistent decrease in accuracy for all models.
The comparison is shown in Figure~\ref{fig:radar_chart_overall}.

\paragraph{ChatGPT over Different Scales} Given that ChatGPT performs exceptionally well among other models, we further analyze ChatGPT as the representative model. 
For data related to other models, please refer to Table~\ref{tab:other_three_model_table} to ~\ref{tab:llama_three_model_table} in the Appendix.We illustrate how ChatGPT performs across different scales from the following aspects:
(1) \textbf{Complexity}: 
As the decoding length for ChatGPT increases from Easy to Hard difficulty, the accuracy decreases consistently across all scales, see Figure~\ref{fig:radar_chart_complexity}.
(2) \textbf{Prompt Method}:
Figure~\ref{fig:radar_chart_prompt} shows that the inclusion of CoT in Zero-shot harms performance across all scales while Few-shot works better for large scales.
(3) \textbf{Language}: 
As both Chinese and English have relative high number system transparency~\cite{johnson-etal-2020-probing}, the gaps between two languages is surprising, see Figure~\ref{fig:radar_chart_language}. 
This partially shows that either training corpus is skewed or numeral conversions knowledge is less transferable across languages.

\subsection{Evaluation of Numerals with Units of Measurement}
In the experiment concerning units of measurement, we adopt the same prompt design as in the previous experiment. To further investigate the impact of unit conversion knowledge on the reasoning capabilities of  LLMs in this experiment, we define \textbf{Few-shot CoT with knowledge} that involves the addition of necessary unit conversion knowledge to the Few-shot CoT framework. For all prompt designs regarding units of measurement, please refer to Table~\ref{tab:unit_of_measurement} in the appendix.

\begin{figure*}[ht]
\centering
\subfloat[The difference in accuracy of the model with and without CoT.\label{fig:unit_add_unit_knowledge_1}]{%
  \includegraphics[width=0.32\textwidth]{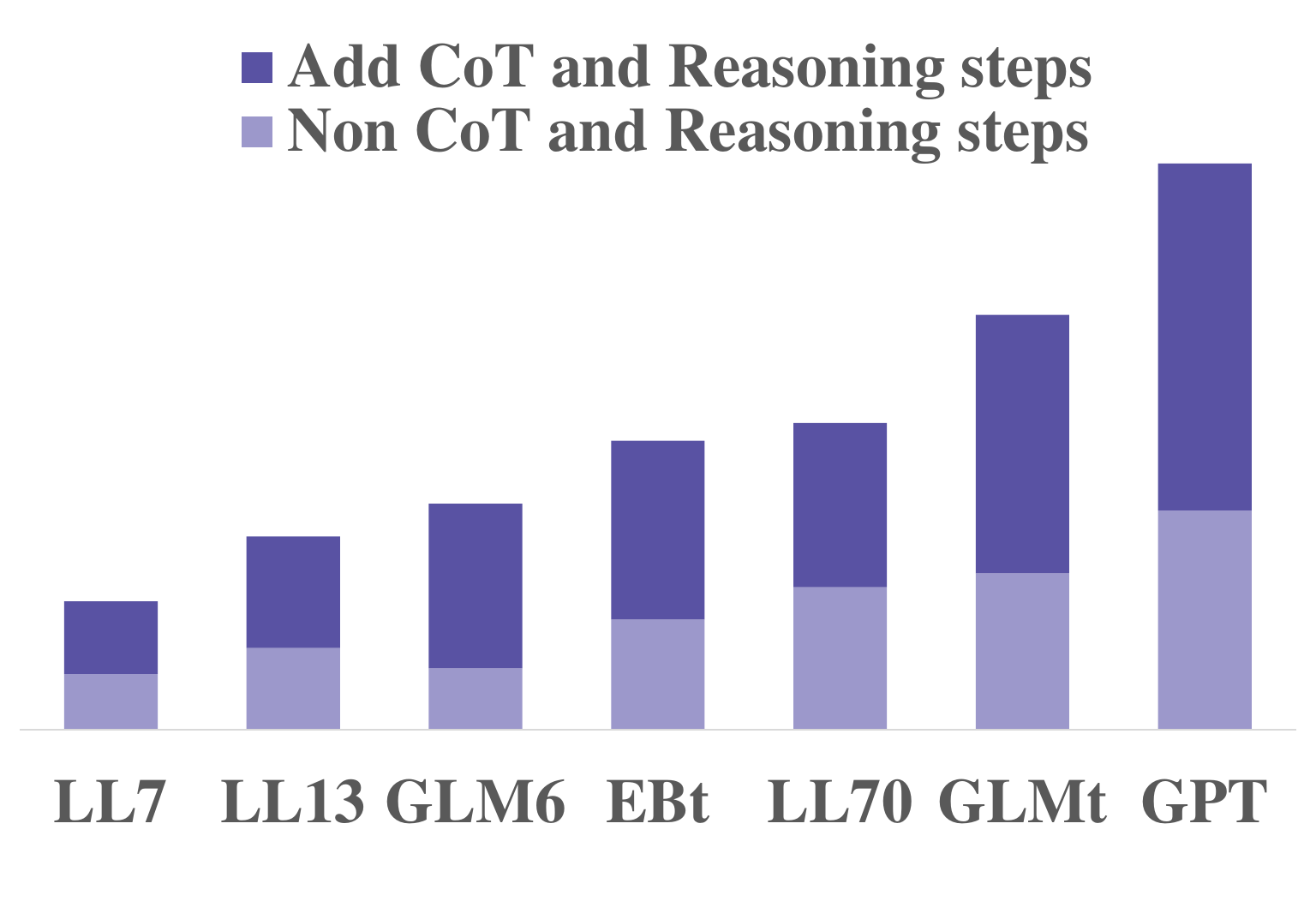}%
}\hfil
\subfloat[The difference in accuracy of the model on datasets of different difficulty levels.
\label{fig:unit_add_unit_knowledge_2}]{%
  \includegraphics[width=0.32\textwidth]{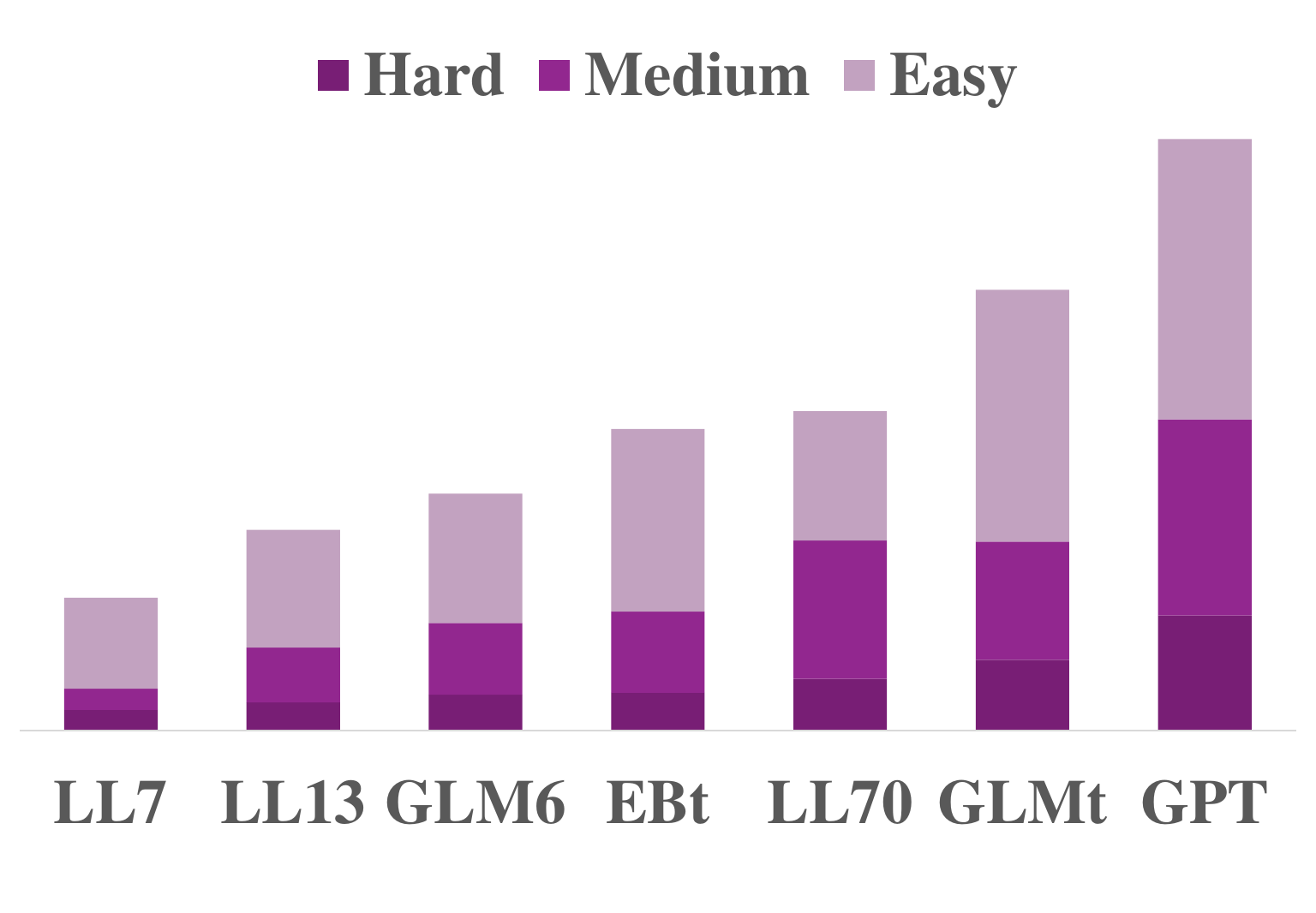}%
}\hfil
\subfloat[The difference in accuracy of the model on datasets of different language.\label{fig:unit_add_unit_knowledge_3}]{%
  \includegraphics[width=0.32\textwidth]{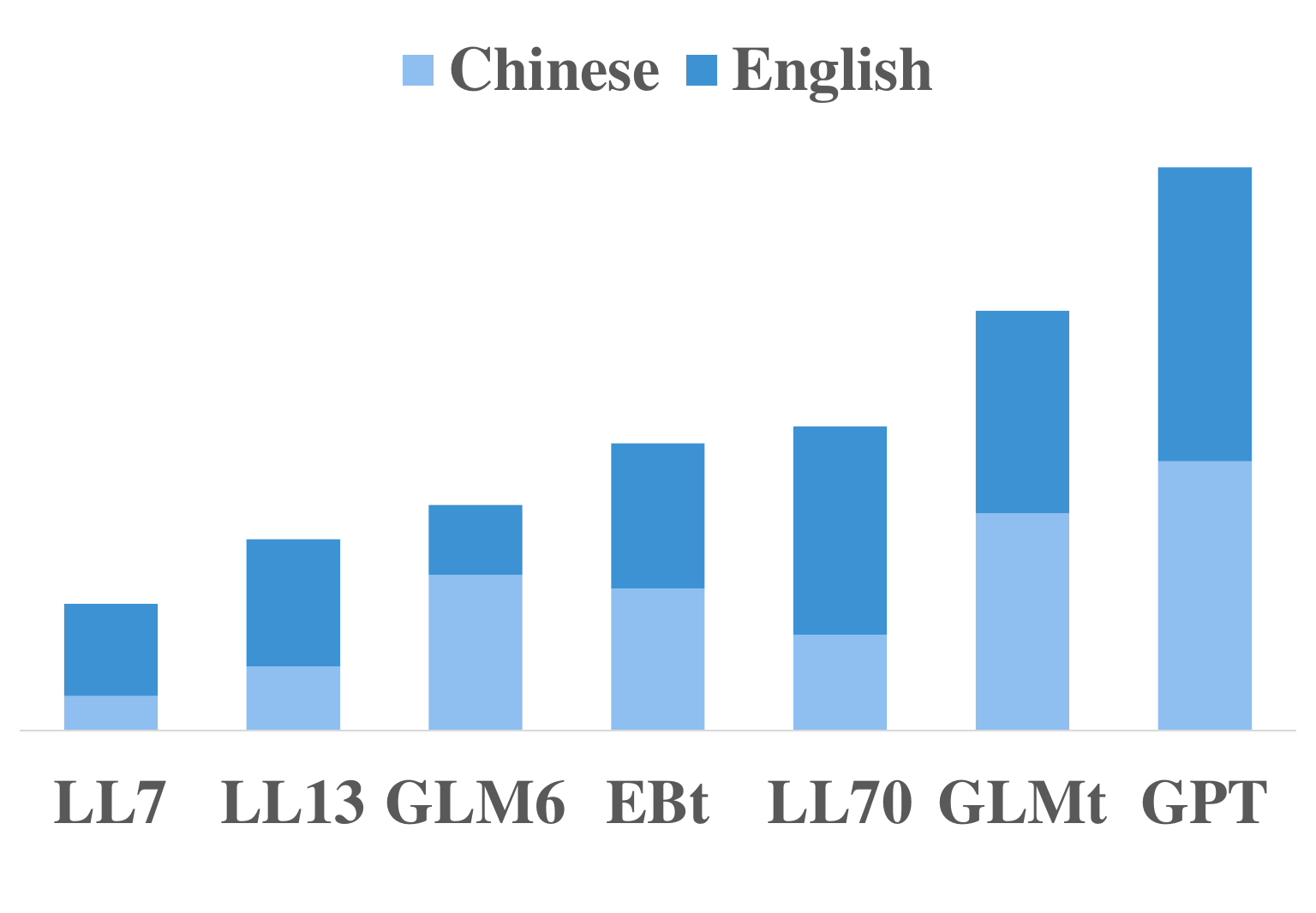}%
}
\caption{Variations in accuracy among LLMs are observed after distinguishing between CoT, difficulty, and language in Units of Measurement problems. Due to space constraints, we use abbreviations here. LL7 represents Llama2-7B, LL13 represents Llama2-13B, LL70 represents Llama2-70B, GLM6 represents ChatGLM2-6B, EBt represents ERNIE-Bot-turbo, GLMt represents ChatGLM-turbo, and GPT represents ChatGPT.} \label{fig:unit_histogram}
\end{figure*}

\begin{table*}[t]
\centering
\includegraphics[width=\linewidth]{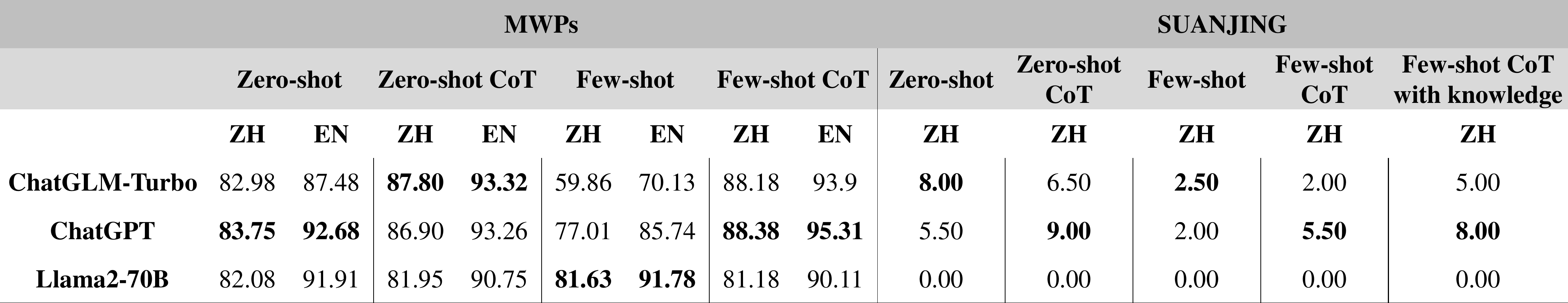}
\caption{Overview of the impact of four prompts and three models on the accuracy of answers in the bilingual MWPs set and the \textsc{SuanJing} set.}
\label{tab:mwp_experience}
\end{table*}

The Table~\ref{tab:unit_experience} is experimental results of the seven models across datasets of three different difficulty levels. The results clearly demonstrate that (1) ChatGPT, compared to other models, consistently exhibits superior performance and reasoning capabilities across all levels of dataset difficulty and in both languages. (2) Unlike the previous Num2words and Words2Num experiments, the introduction of CoT and reasoning steps in this experiment significantly enhances the success rate of LLMs in accurately generating answers.

To delve into more specific information, we categorize the experimental data and create three bar graphs as depicted in the Figure~\ref{fig:unit_histogram}. Figure~\ref{fig:unit_add_unit_knowledge_1} illustrates that the introduction of CoT and reasoning steps led to a noticeable improvement in the accuracy of each model when handling units of measurement tasks, Figure~\ref{fig:unit_add_unit_knowledge_2} shows that as the difficulty of the questions increased, the accuracy of each model in dealing with units of measurement tasks decreased correspondingly. Figure~\ref{fig:unit_add_unit_knowledge_3} indicates that the models exhibit roughly the same accuracy in handling tasks in both Chinese and English, even in the case of ChatGPT.

\subsection{Comparisons over MWPs and \textsc{SuanJing}}

In this section, we employ three state-of-the-art models, ChatGPT, ChatGLM-Turbo and Llama2-70B, to evaluate the performance of LLMs on MWPs and \textsc{SuanJing}. We select 100 questions from \textsc{SuanJing} that share the same operators and complexity level as MWPs.
Additionally, \textsc{SuanJing} poses extra challenges due to the textual format of numbers and the requirement for unit conversions during the reasoning process when necessary.

The experimental results for MWPs are presented in the left part of Table~\ref{tab:mwp_experience}, which clearly indicate the following:
(1) LLMs can handle both English and Chinese questions very effectively.
(2) The incorporation of CoT significantly enhances the accuracy of LLMs on MWPs.

Contrary to this, the experimental results for \textsc{SuanJing}, as shown in the right part of Table~\ref{tab:mwp_experience}, demonstrate a substantial decline in accuracy. 
Furthermore, the addition of CoT does not yield any improvement and, in fact, seems to deteriorate the performance.

\subsection{Analysis of Errors from LLMs}

\begin{figure*}[t]
\centering
\subfloat[Errors in numeral conversion experiments.\label{fig:error_1}]{%
  \includegraphics[width=0.5\textwidth]{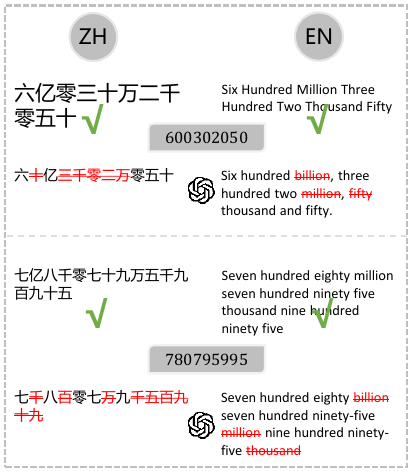}%
}\hfil
\subfloat[Errors in units of measurement experiments.\label{fig:error_2}]{%
  \includegraphics[width=0.5\textwidth]{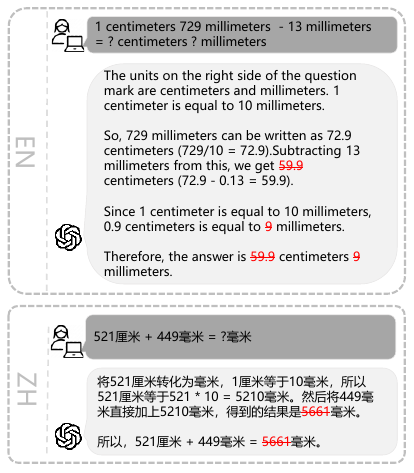}%
}\hfil

\caption{Common errors in Numeral and Units of Measurement experiments from LLMs.}
\label{fig:errors_2}
\end{figure*}

\begin{figure}[t]
\includegraphics[width=\linewidth]{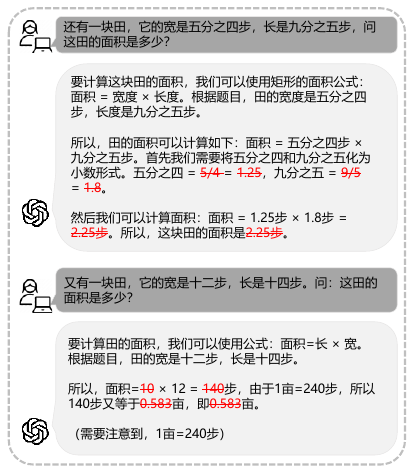}
\caption{Errors in \textsc{SuanJing} experiments.}
\label{fig:error_3}
\end{figure}

In the numeral conversions experiments, the majority of errors are primarily concentrated on LLMs' inability to correctly identify the magnitudes of individual numbers. As illustrated in the figure~\ref{fig:error_1}, the number 6, originally representing hundred million in Arabic numerals, is incorrectly identified as billion. Additionally, errors arise from LLMs' mishandling of the digit zero, including failure to recognize its significance and inability to accurately restore the quantity and position of zero in the numerical context. To enhance the accuracy of LLMs in such tasks, future improvements could focus on refining LLMs' ability to recognize the length and magnitude of numbers.

Our experiments also demonstrated that CoT did not work in the numeral conversion experiments.
LLMs achieved significantly higher accuracy rates on the \textit{easy} dataset, which was of comparable scale to the \textit{medium} dataset but required shorter answer lengths.
This discrepancy highlights two main challenges that LLMs face in numerical reasoning. First, the linguistic nature of input text makes it difficult for LLMs to understand numerical data. Second, the flexibility and complexity of the answers increase the likelihood of errors in longer outputs.
Given that CoT primarily enhances performance on complex inference tasks rather than simple ones, its application to simpler tasks such as Num2Words and Words2Num increases the length of the generated text, thereby diminishing LLMs accuracy.

In the units of measurement experiment, the majority of errors primarily stem from LLMs' failure to correctly recognize the conversion magnitude relationship when multiple units are involved. 
As depicted in the figure~\ref{fig:error_2}, there exists a tenfold progressive relationship between decimeters and centimeters, yet LLMs overlook the magnitude relationship inherent in textual units. Introducing CoT significantly mitigates the occurrence of such errors but still requires further refinement. Additionally, even when LLMs have correctly grasped the magnitude relationship inherent in textual units, errors may still occur during the calculation process. To enhance the accuracy of LLMs in such tasks, efforts could be directed towards improving LLMs' recognition of textual units and the magnitude relationships between units.

In the \textsc{SuanJing} experiment, LLMs face more comprehensive problem-solving tasks. As depicted in the figure~\ref{fig:error_3}, LLMs encounter errors in handling units in \textsc{SuanJing} problems, as some ancient units are extremely rare in contemporary society, making it difficult for LLMs to correctly understand \textsc{SuanJing} problems. This long tail problem can be addressed by introducing external knowledge in the prompt, thereby enabling LLMs to have a chance of correctly handling \textsc{SuanJing} problems. However, LLMs frequently make errors in recognizing numbers and performing numerical calculations, especially in the recognition and computation of more challenging fractions and decimals. Consequently, even if LLMs can correctly utilize the external prompt-introduced knowledge of ancient units, their accuracy remains relatively low. Due to the extensive use of fractions described in Classical Chinese in \textsc{SuanJing}, LLMs need to undergo multiple Hard-level Words2Num tasks before answering questions, significantly reducing the accuracy of \textsc{SuanJing} experiments.

In the MWPs experiment, the majority of errors are similar to those in the units of measurement experiment, as the MWPs experiment can be considered a natural language version of the units of measurement experiment to some extent. Furthermore, \textsc{SuanJing} can be seen as a more challenging version of MWPs, hence many errors observed in the preceding experiments are also frequent in \textsc{SuanJing}. To improve the accuracy of LLMs in such tasks, besides focusing on the improvement directions of units of measurement experiments, attention should also be given to the performance of LLMs on long tail problems.

\section{Conclusion}
We investigate the performance of various LLMs on tasks involving numeral conversions and units of measurement in both Chinese and English languages. Additionally, we explore the capability boundaries of LLMs by introducing CoT and external knowledge. Based on a series of experiments, the conclusions are as follows: 

\begin{enumerate}
    \item There is a noticeable performance gap between Chinese LLMs and top-tier models like ChatGPT. 
    \item The same large language model exhibits varying levels of performance facing problems in different languages. 
    \item Despite the introduction of external knowledge and CoT, LLMs still struggle to effectively handle comprehensive problems involving numeral conversions and units of measurement.
\end{enumerate}

\section*{Limitations}
In this paper, we introduce datasets to investigate whether LLMs can process numeral conversions and units of measurement like humans, despite certain limitations.

Firstly, LLMs' responses exhibit randomness, and we do not conduct repeated trials of the same question with the same model. Conducting repeated trials could reduce randomness and yield more precise accuracy estimates.
Secondly, Our experiments involve only seven types of models. Incorporating commercial models such as GPT-4 and Claude might provide a more representative performance curve. 

Future work could diversify experimental data by incorporating datasets from professional backgrounds like financial accounting, aiding in exploring the performance boundaries of LLMs.

\section*{Acknowledgements}
This work was partially supported by the National Key Research and Development Program of China (2022YFF0902100),  China Postdoctoral Science
Foundation~(2023M733654), National Natural Science Foundation of China (62376262),  Guangdong Basic and Applied Basic Research Foundation~(2023A1515110496), the Natural Science Foundation of Guangdong Province of China (2024A1515030166), Shenzhen Science and Technology Innovation Program (KQTD20190929172835662), Shenzhen Basic Research Foundation (JCYJ20210324115614039).

\bibliography{anthology,custom}
\bibliographystyle{acl_natbib}

\appendix

\section{\textsc{SuanJing} Dataset}
\label{sec:appendix}

\begin{table}[!htp]\centering
\small
\begin{tabular}{cp{3.5cm}r}\toprule
\multicolumn{2}{c}{Title} &Count \\\midrule
《周髀算经》& \textit{Zhou Shadow Mathematical Classic} & - \\
《九章算术》& \textit{The Nine Chapters on the Mathematical Art} &246 \\
《海岛算经》& \textit{The Sea Island Mathematical Classic} &9 \\
《孙子算经》& \textit{The Mathematical Classic of Sun Zi} &65 \\
《张邱建算经》& \textit{The Mathematical Classic of Zhang Qiujian} &92 \\
《五曹算经》& \textit{Computational Canon of the Five Administrations} &68 \\
《夏侯阳算经》&\textit{The Mathematical Classic of Xiahou Yang} &82 \\
《五经算术》 &\textit{Computational Prescriptions of the Five Classics} & \\
《缉古算经》 &\textit{Continuation of Ancient Mathematical Classic} &20 \\

《缀术》 &\textit{Method of Interpolation} &- \\
《益古演段》&\textit{Old Mathematics in Expanded Sections} &64 \\
《数学九章》&\textit{The Mathematical Treatise in Nine Sections} &80 \\\midrule
Total& & 726 \\
\bottomrule
\end{tabular}
\caption{Statistics for math word problems extracted from ancient Chinese mathematics classics.}\label{tab:suanjing}
\end{table}

To facilitate the evaluation of reasoning integrating all sub-procedures, we need a dataset with challenges discussed above.
We construct \textsc{SuanJing}~(算经) by extracting and annotating math word problems from a collection of ancient Chinese algorithmic books.

Although grammars and lexicons of the Chinese language endure great changes in history, the numeral systems and units of measurement are reserved and still used in daily life.
Especially in formal documents and statements of financial institutions, the representation of numerals are required to be written in both traditional style\footnote{\url{https://en.wikipedia.org/w/index.php?title=Chinese_numerals}} and Hindu–Arabic style in order to avoid subsequent manipulations.
This presents us a great opportunity to reuse ancient math word problems and look closely at how numeral systems and units of measurement affect reasoning steps of LLMs. 

Problems in \textsc{SuanJing} are collected from ancient Chinese mathematical classics.
Since Tang Dynasty~(唐朝), Mingsuan~(明算, comprehend of arithmetic) has been an important subject in Keju~(科举,  imperial examinations) for bureaucrats selection.
Mathematician Li Chunfeng\footnote{\url{https://en.wikipedia.org/w/index.php?title=Li_Chunfeng}} edited The Ten Computational Canons\footnote{\url{https://en.wikipedia.org/w/index.php?title=Ten_Computational_Canons}}, which was a collection of ten Chinese mathematical works.
We additionally add \textit{Old Mathematics in Expanded Sections)} and \textit{The Mathematical Treatise in Nine Sections} to \textsc{SuanJing}.
The full list of classics and extracted problem counts are shown in Table~\ref{tab:suanjing}.
 
\begin{table*}[!hpt]
\centering
\includegraphics[width=\linewidth]{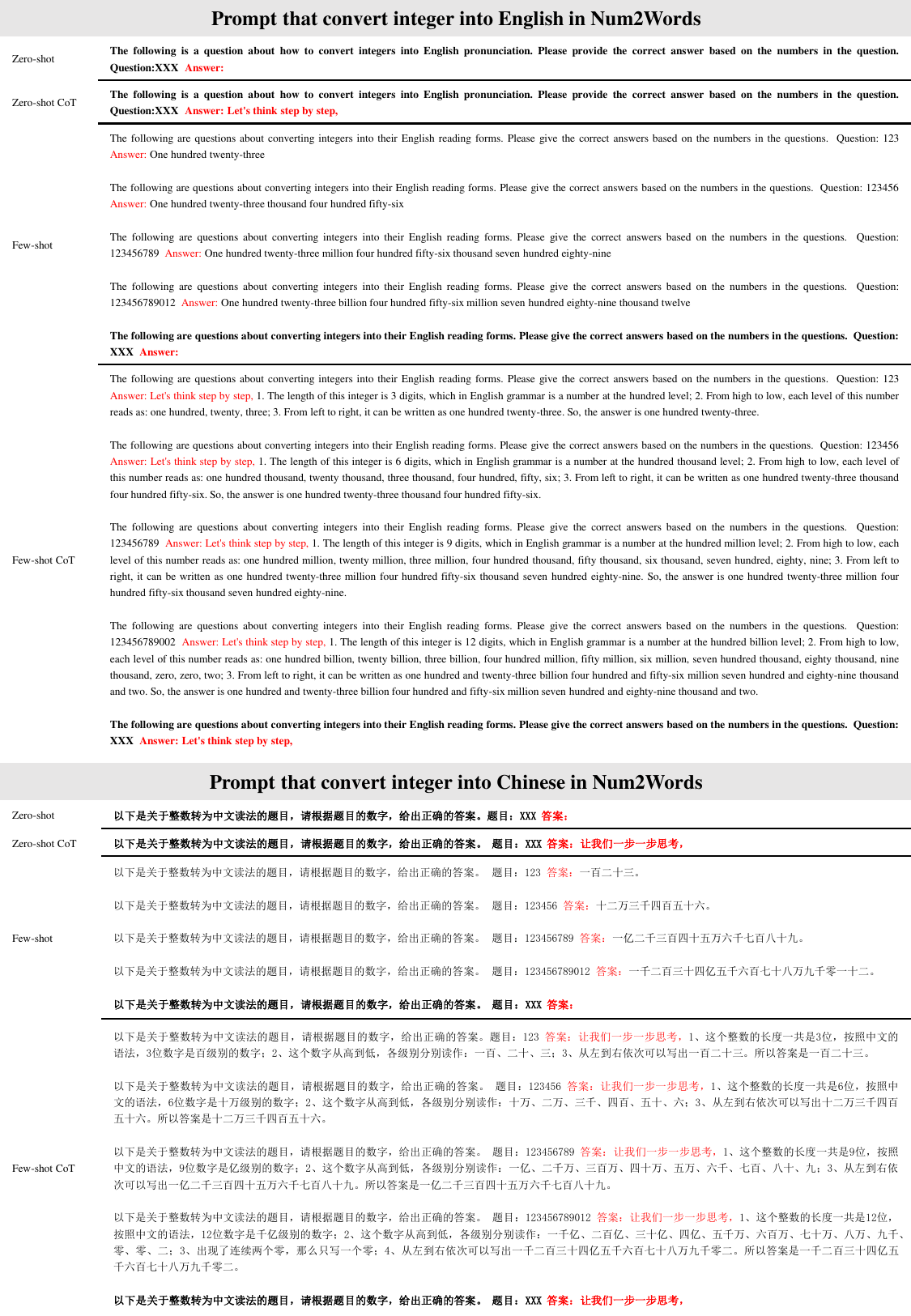}
\caption{Prompt that convert integers to English and Chinese in Num2Words task, ‘XXX’ is a word or numerical question in the dataset.} 
\label{tab:prompt_num2word-1}
\end{table*}

\begin{table*}[!hpt]
\centering
\includegraphics[width=\linewidth]{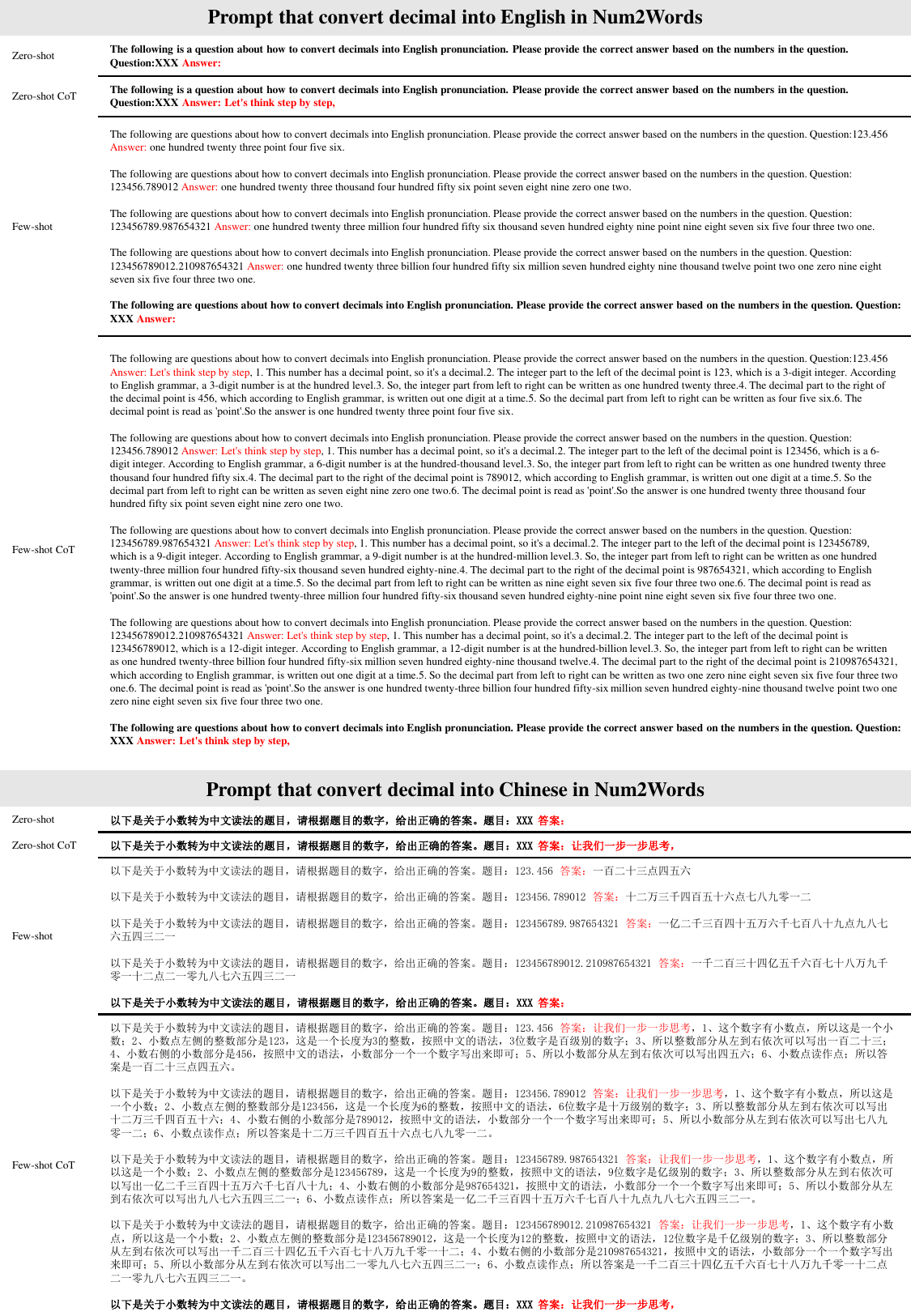}
\caption{Prompt that convert decimals to English and Chinese in Num2Words task, ‘XXX’ is a word or numerical question in the dataset.} 
\label{tab:prompt_num2word-2}
\end{table*}

\begin{table*}[!hpt]
\centering
\includegraphics[width=\linewidth]{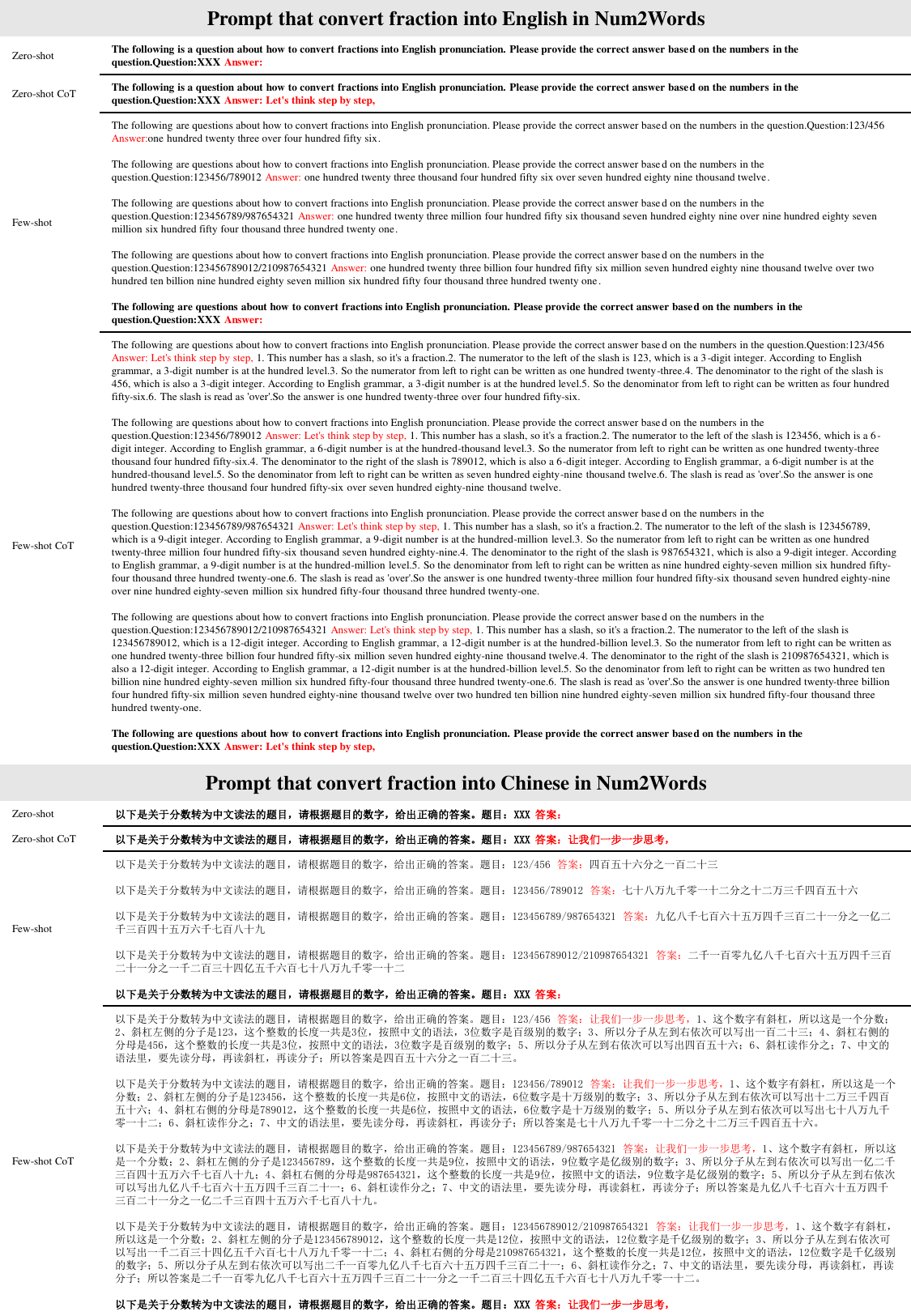}
\caption{Prompt that convert fractions to English and Chinese in Num2Words task, ‘XXX’ is a word or numerical question in the dataset.} 
\label{tab:prompt_num2word-3}
\end{table*}

\begin{table*}[!hpt]
\centering
\includegraphics[width=\linewidth]{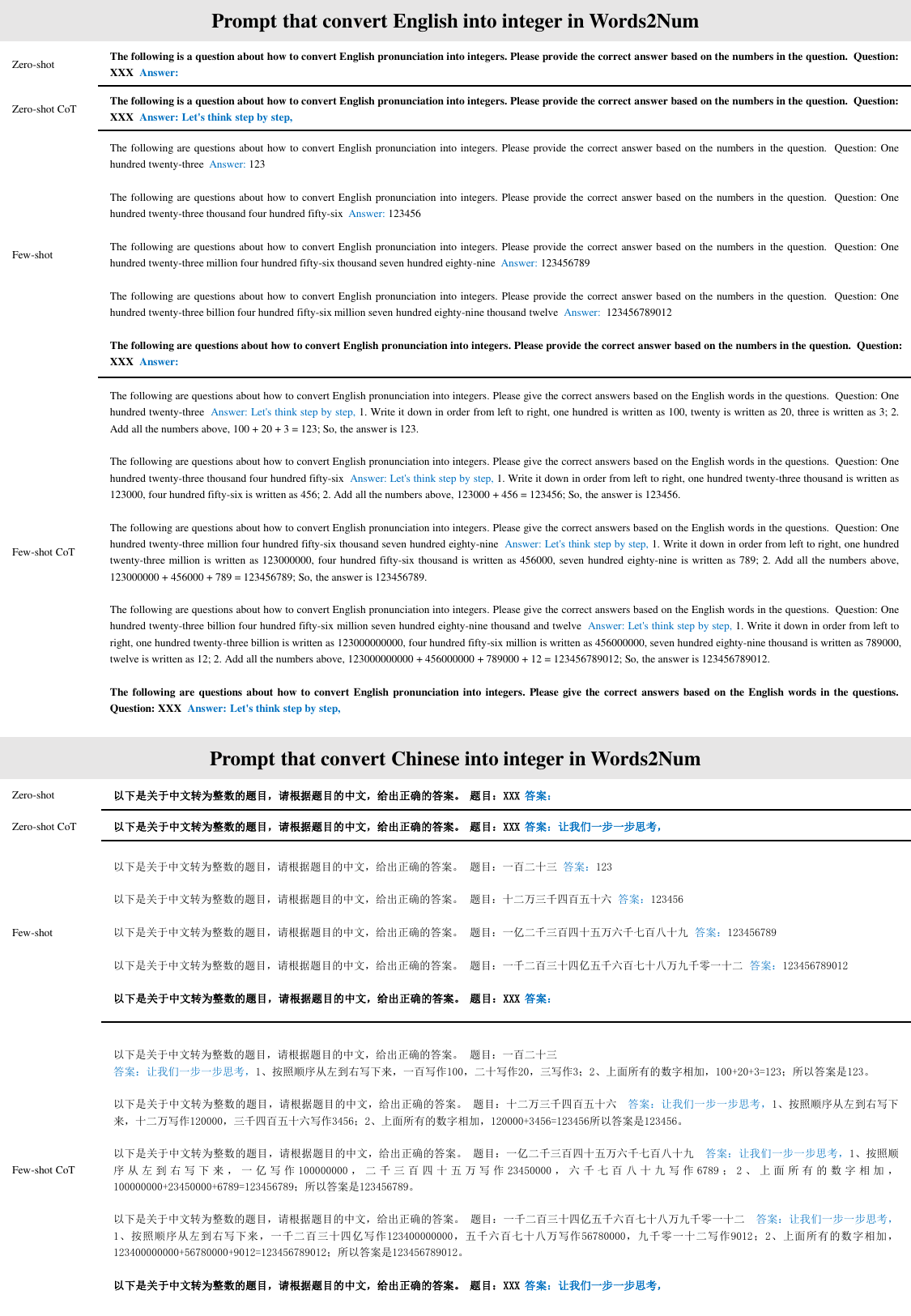}
\caption{Prompt that convert English and Chinese to integer in Words2Num task, ‘XXX’ is a word or numerical question in the dataset.} 
\label{tab:prompt_word2num-1}
\end{table*}

\begin{table*}[!hpt]
\centering
\includegraphics[width=\linewidth]{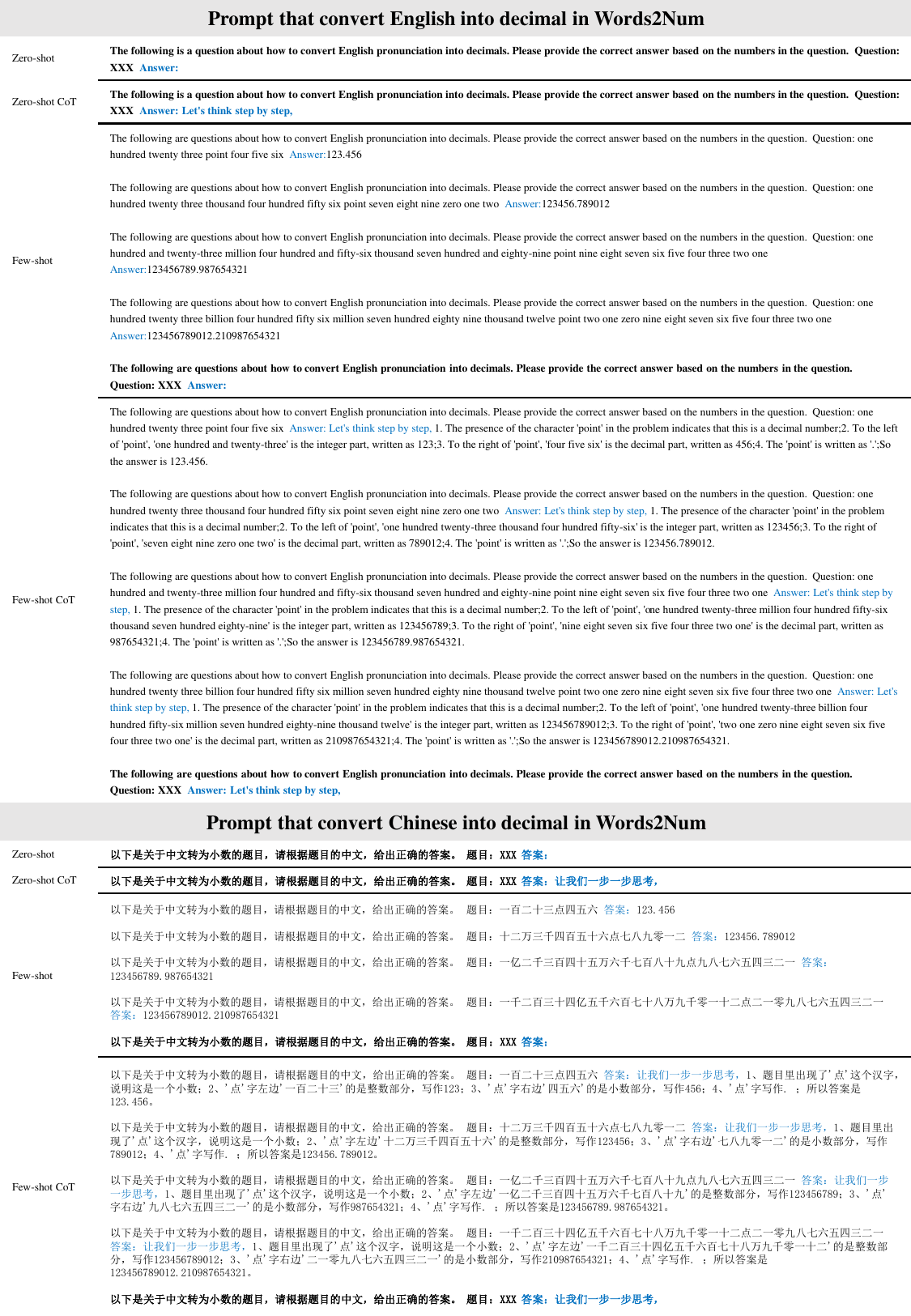}
\caption{Prompt that convert English and Chinese to decimal in Words2Num task, ‘XXX’ is a word or numerical question in the dataset.} 
\label{tab:prompt_word2num-2}
\end{table*}

\begin{table*}[!hpt]
\centering
\includegraphics[width=\linewidth]{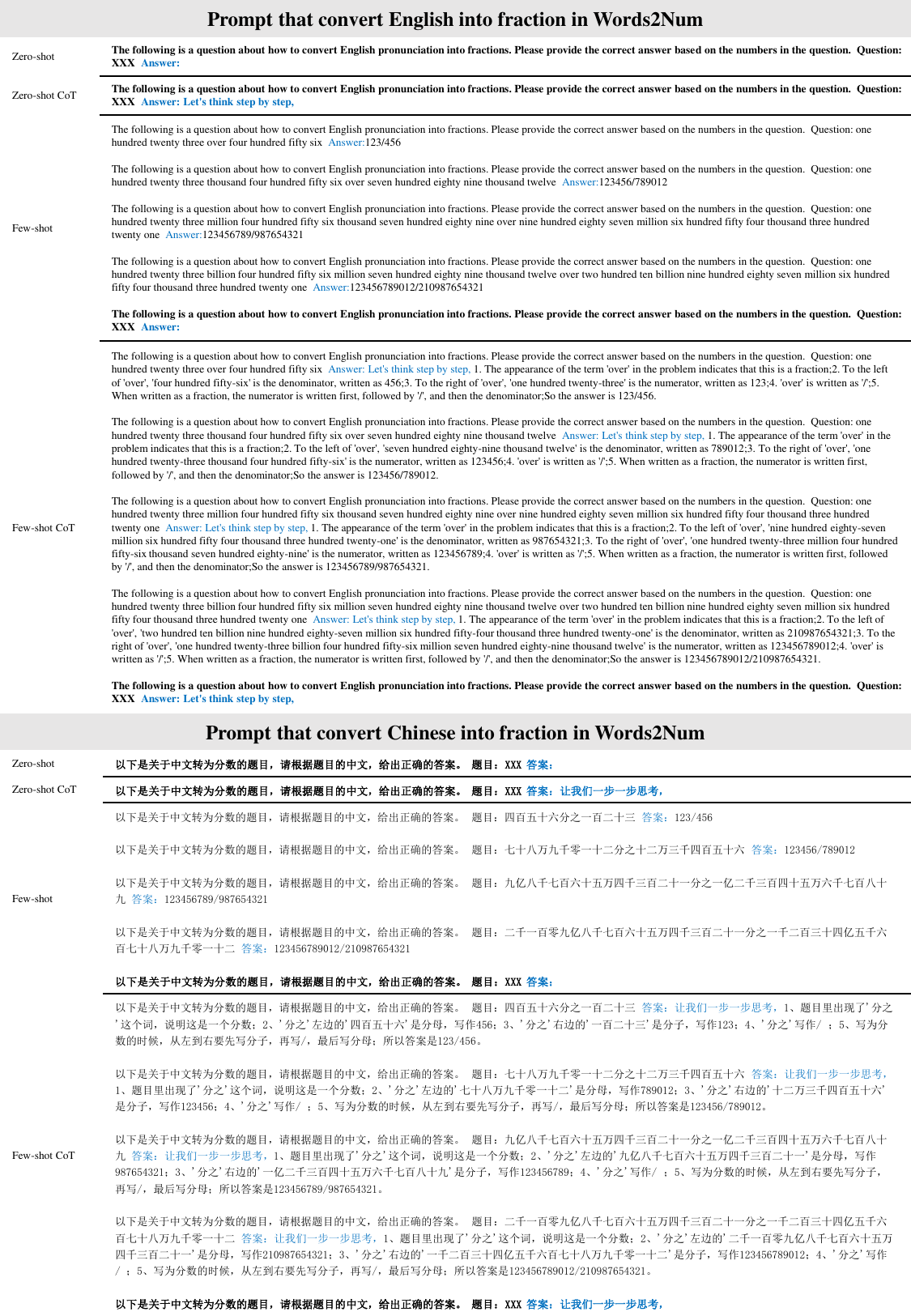}
\caption{Prompt that convert English and Chinese to fraction in Words2Num task, ‘XXX’ is a word or numerical question in the dataset.} 
\label{tab:prompt_word2num-3}
\end{table*}

\begin{table*}[!hpt]
\centering
\includegraphics[width=\linewidth]{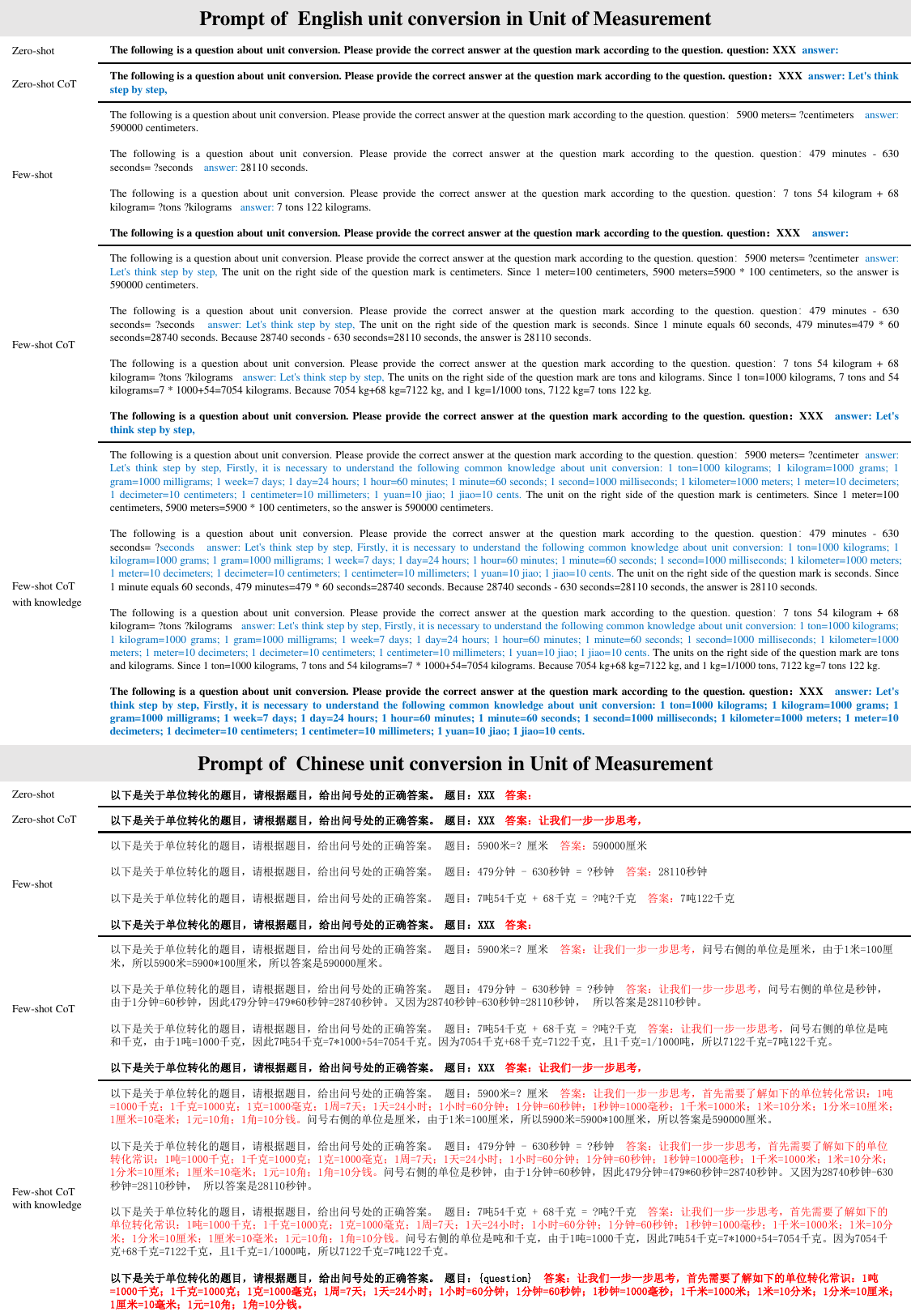}
\caption{Prompt of English and Chinese unit conversion in Units of Measurement task, ‘XXX’ is a word or numerical question in the dataset.} 
\label{tab:unit_of_measurement}
\end{table*}

\begin{table*}[!hpt]
\centering
\includegraphics[width=\linewidth]{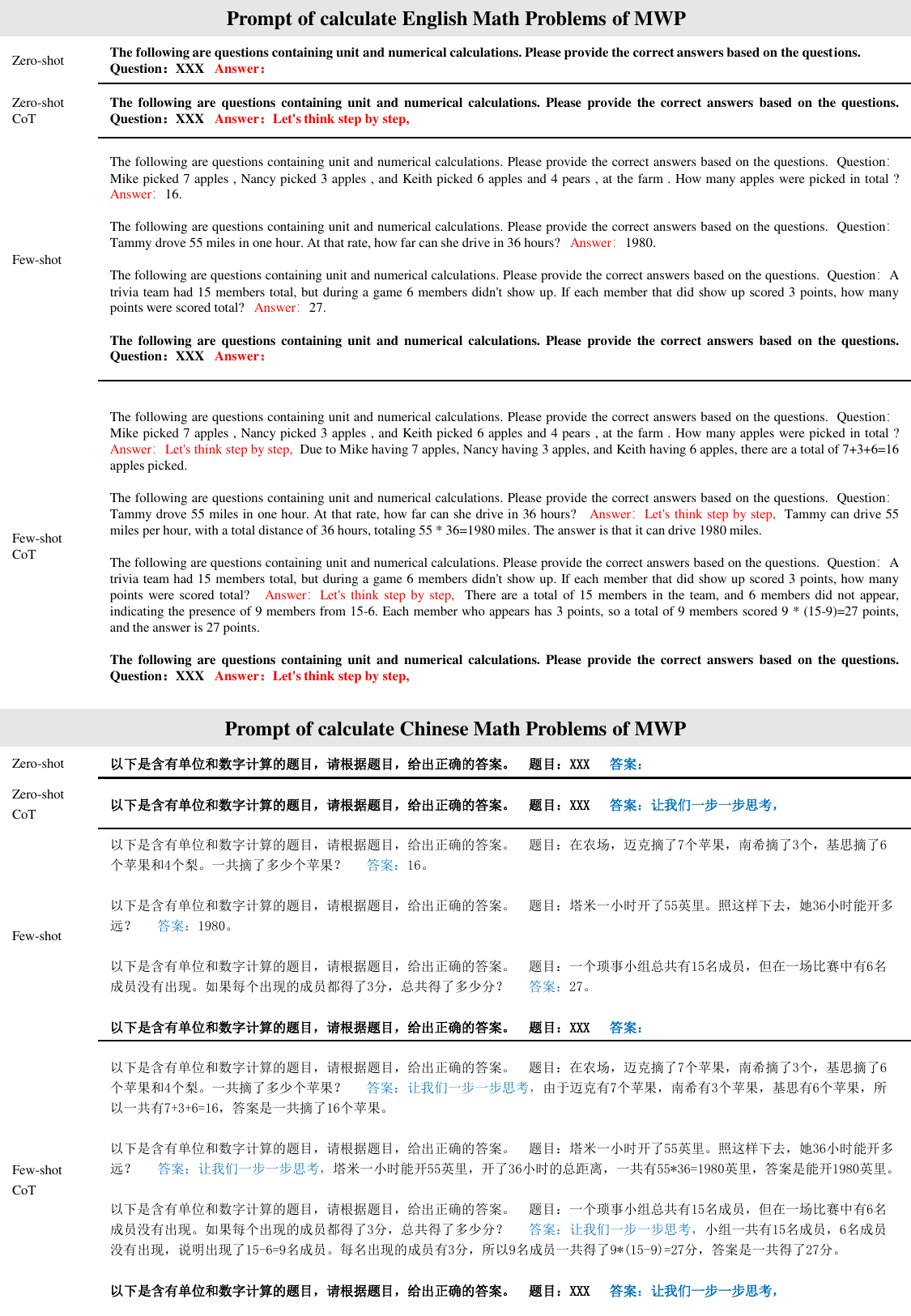}
\caption{Prompt of calculate English and Chinese math problems of MWPs, ‘XXX’ is a question in the dataset.} 
\label{tab:prompt_mwp}
\end{table*}

\begin{table*}[!hpt]
\centering
\includegraphics[width=\linewidth]{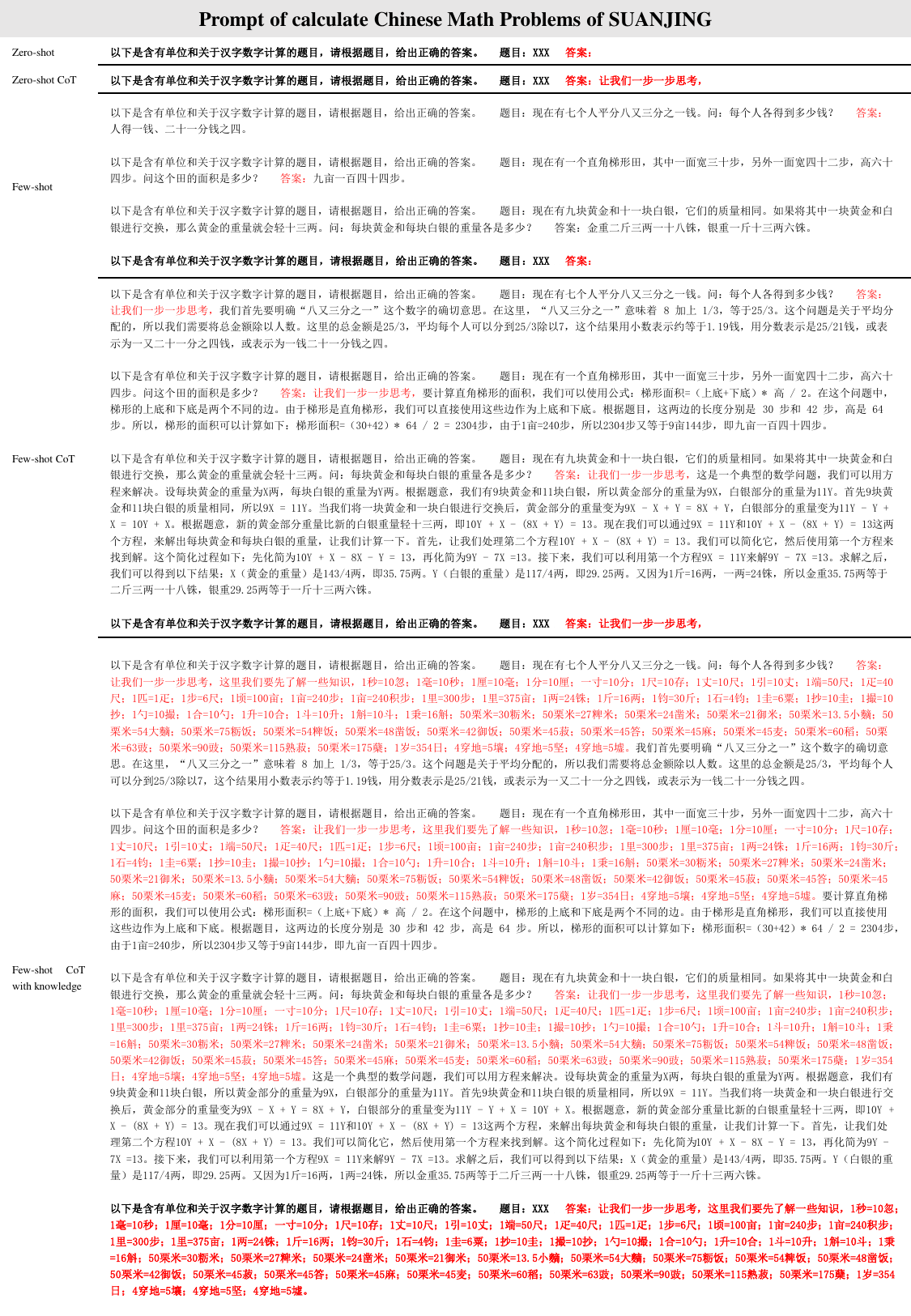}
\caption{Prompt of calculate the modern version of mathematical problems of \textsc{SuanJing}, ‘XXX’ is a question in the dataset.} 
\label{tab:prompt_suanjing}
\end{table*}

\begin{table*}[!hpt]
\centering
\includegraphics[width=\linewidth]{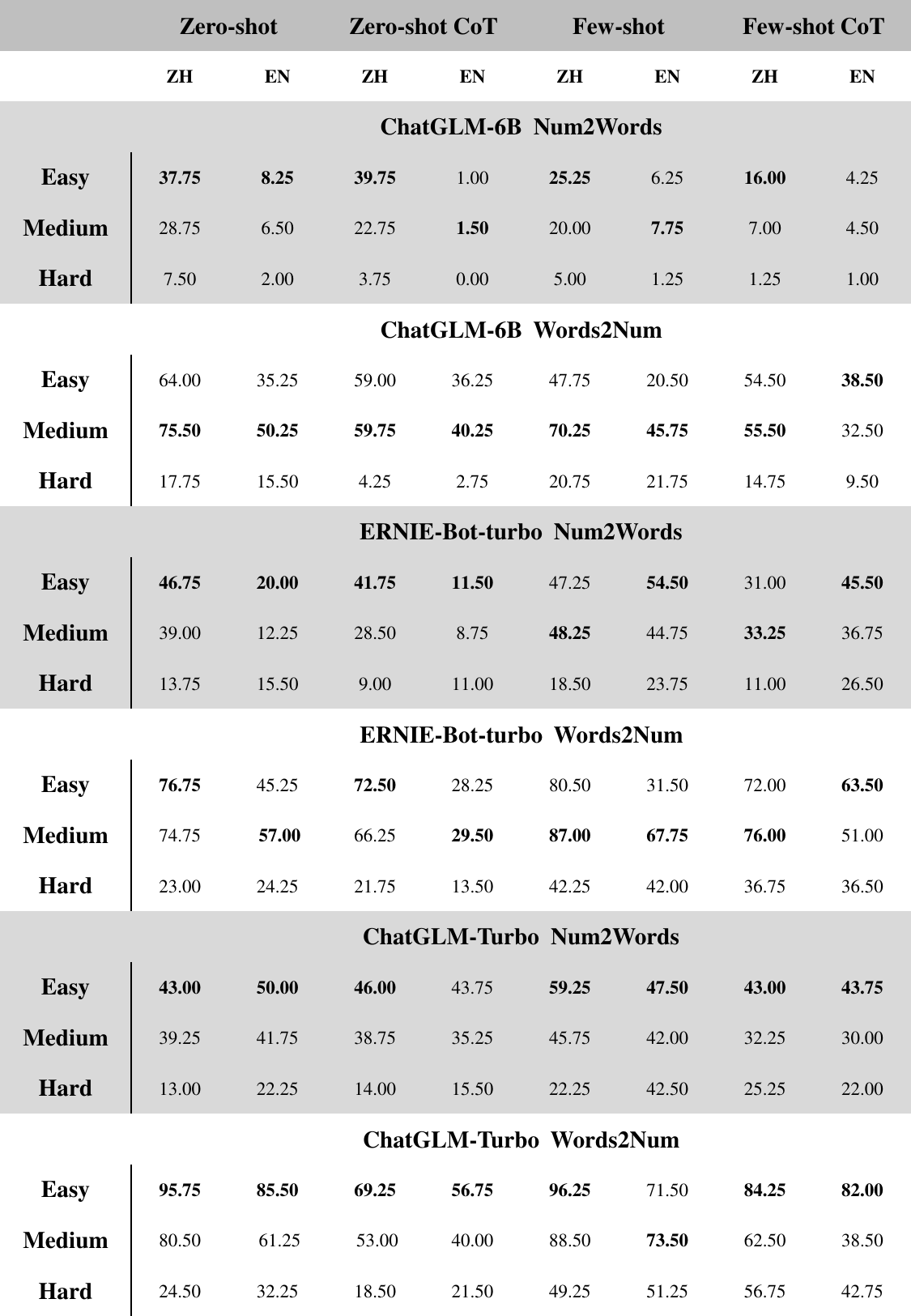}
\caption{The accuracy performance of models ChatGLM-6B, ERNIE-Bot-turbo, and ChatGLM-Turbo in different difficulty levels of Num2words and Words2Num tasks.} 
\label{tab:other_three_model_table}
\end{table*}

\begin{table*}[!hpt]
\centering
\includegraphics[width=\linewidth]{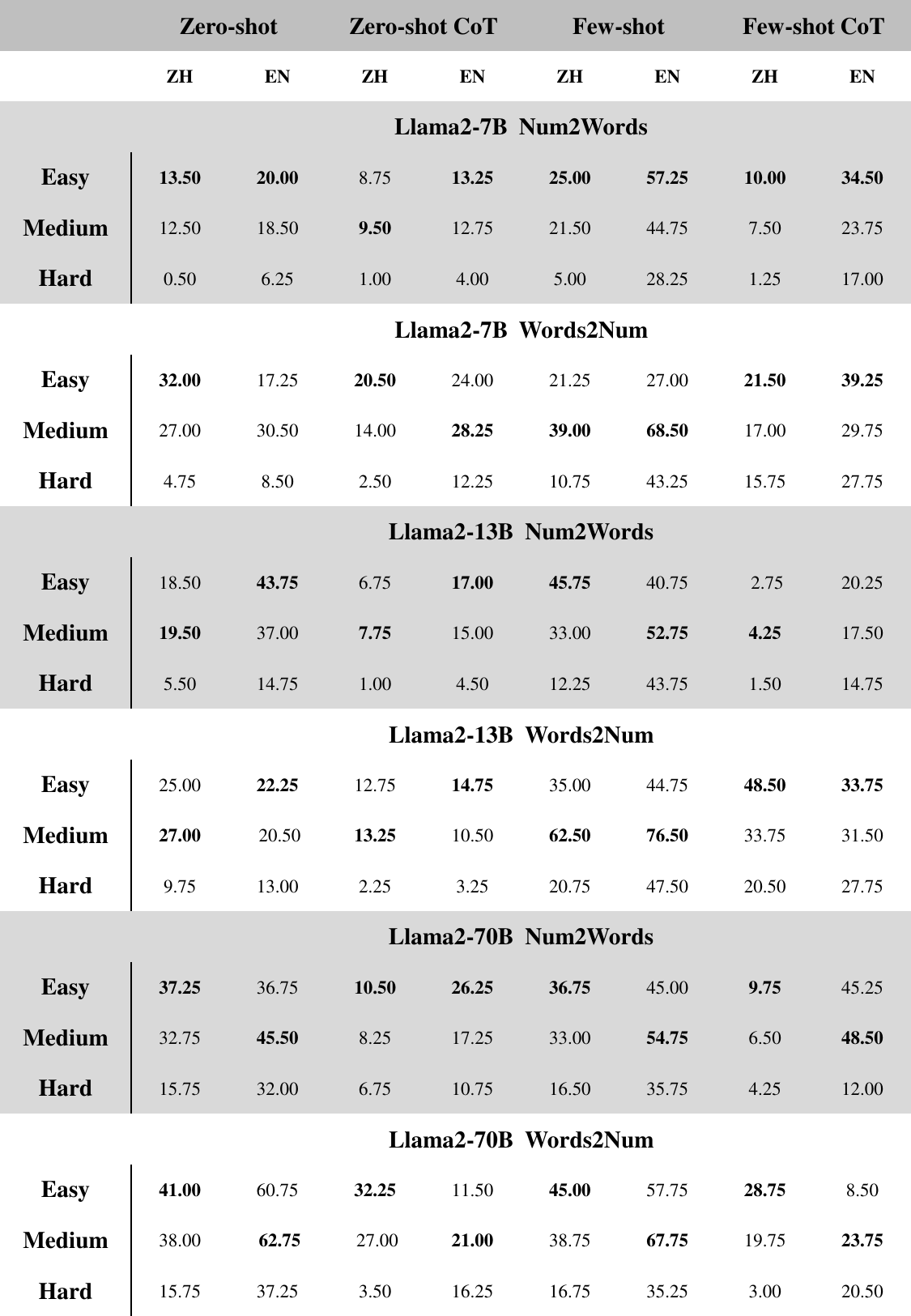}
\caption{The accuracy performance of models Llama2-7B, Llama2-13B and Llama2-70B in different difficulty levels of Num2words and Words2Num tasks.} 
\label{tab:llama_three_model_table}
\end{table*}

\end{CJK*}

\end{document}